\pdfoutput=1

\documentclass[11pt,dvipsnames]{article}

\usepackage{ACL2023}

\usepackage{times}
\usepackage{latexsym}

\usepackage[T1]{fontenc}

\usepackage[utf8]{inputenc}

\usepackage{microtype}

\usepackage{inconsolata}
\usepackage{amsmath}

\newcounter{notecounter}
\newcommand{\enotesoff}{\long\gdef\enote##1##2{}}
\newcommand{\enoteson}{\long\gdef\enote##1##2{{
			\stepcounter{notecounter}
			\large\bf
			\hspace{100cm}\arabic{notecounter} $<<<$ ##1: ##2
			$>>>$\hspace{1cm}}}}
\enoteson
\enotesoff

\newcommand{\xlmr}{XLM\mbox{-}R}

\usepackage{colortbl}
\usepackage{arydshln} 

\definecolor{existing-1}{HTML}{FCAC5D}
\definecolor{existing-2}{HTML}{FCB75D}
\definecolor{existing-3}{HTML}{FCC75D}
\definecolor{existing-4}{HTML}{FCD45D}
\definecolor{existing-5}{HTML}{FCE45D}
\definecolor{existing-6}{HTML}{FCF45D}

\definecolor{new-1}{HTML}{9DA4FF}
\definecolor{new-2}{HTML}{A5A8FF}
\definecolor{new-3}{HTML}{C4BAFF}
\definecolor{new-4}{HTML}{DFC2FF}
\definecolor{new-5}{HTML}{EBCCFF}
\definecolor{new-6}{HTML}{F6D5FF}

\title{Understanding Cross-Lingual Alignment---A Survey}

\author{
Katharina Hämmerl$^{1,2}$ \and
  Jindřich Libovický$^{3}$  \and
  Alexander Fraser$^{2,4}$ \\
  \\
$^1$Center for Information and Language Processing, LMU Munich, Germany \\
\texttt{haemmerl[at]cis.lmu.de}\\
$^2$Munich Centre for Machine Learning (MCML), Germany \\
$^3$Faculty of Mathematics and Physics, Charles University, Czech Republic \\
$^4$Technical University of Munich, Germany
}

\begin{document}
\maketitle
\begin{abstract}
Cross-lingual alignment, 
the meaningful similarity of representations across languages
in multilingual language models, has been an active field of research in recent years. 
We survey the literature of techniques to improve cross-lingual alignment, providing a taxonomy of methods and summarising insights from throughout the field.
We present
different understandings of cross-lingual alignment and their limitations.
We provide a qualitative summary of results from
a large number of 
surveyed papers.
Finally, we discuss how these insights may be applied not only to encoder models, where this topic has been heavily studied, but also to encoder-decoder or even decoder-only models, and argue that 
an effective trade-off between language-neutral and language-specific information
is key.
\end{abstract}

\section{Introduction}

Zero-shot cross-lingual transfer using highly multilingual models has been an active subset of multilingual NLP research.
In tasks like sentence classification, sequence labelling, or sentence retrieval, all of which rely on encoder representations, \textit{cross-lingual alignment} of those representations is an underlying assumption for zero-shot cross-lingual transfer.
Once the model has learned to do, e.g., a classification task in a source language, then if the representation of a target language is ``aligned'' to that of the source language, the model should also be able to classify target language items.

As we define it, cross-lingual alignment means that words or sentences with similar semantics: 1) 
 Are more similar in the representation space than words or sentences with dissimilar semantics. This way of looking at alignment can be defined in ``weak'' or ``strong'' terms (see \S~\ref{sec:understanding-alignment}).
 2)
Allow a prediction head trained on a source language to recognise the relevant patterns in the target language.
    We argue that this is related to
    the first view but, importantly, is 
    more nuanced.
    This allows us to discuss the literature in a new way.

These 
two 
criteria are not guaranteed to be fulfilled through unsupervised pre-training, thus
motivating various
efforts to improve cross-lingual alignment.
We survey important papers in this area between 2019 and 2023.
These works propose new training objectives, new pre-trained models, contrastive fine-tuning, or post-hoc adjustments of the embedding space.
The vast majority of the methods were developed for and applied to multilingual encoder models, chiefly \xlmr\ and mBERT.

In this paper, we provide
\begin{itemize}
    
\item a definition of cross-lingual alignment under 
    both
    views, how it is 
    typically
    measured---and weaknesses of these measurements---, and an important critique of ``strong'' alignment as stated by the first view (\S~\ref{sec:understanding-alignment});
\item a taxonomy of methods to increase alignment taken from a comprehensive survey of the literature
    (\S~\ref{sec:increase-alignment});
\item a summary of findings and where the field currently stands (\S~\ref{sec:state-of-the-field});
\item and finally a discussion of ongoing and future research in this area as the field focuses more on
    generative models (\S~\ref{sec:multiling-gen}). As we discuss, generative models pose new challenges for cross-lingual alignment, since we need to trade-off language-neutral with language-specific elements in new ways.
\end{itemize}

\section{Cross-Lingual Alignment}\label{sec:understanding-alignment}

\subsection{Definitions}\label{subsec:definitions}

``Alignment'' is an overloaded term in NLP, referring to word alignment in machine translation \citep{och-etal-1999-improved}, or to desirable model behaviour in chatbot training \citep{ouyang2022training}, or as in our case, 
to the \textit{meaningful similarity of multilingual representations across languages}.
``Cross-lingual alignment'' in this sense was used
in static word embeddings, and can be applied to contextual models as well.
We define the two main views below.

\paragraph{View I.} Similar meanings across languages have more similar representations than dissimilar meanings do.
This view is particularly salient for tasks using, e.g., cosine similarity of representations.
It relies on the embeddings as a whole being distributed ``well''.
There are ``weak'' and ``strong'' definitions of cross-lingual alignment that focus on this view (\citealp[cf.][]{roy-etal-2020-lareqa, gaschi2022multilingual, abulkhanov-etal-2023-lapca}, inter alia).

Weak alignment requires only that the nearest \textit{target-language} neighbour of a word or sentence representation be its target-language translation.
If $(u_i,v_i)$ is a pair of encoder representations corresponding to a pair of equivalent words (or sentences) from the languages $L1$ and $L2$, $s$ is a similarity function, and $N$ is the number of translation pairs in a corpus:

\begin{equation}\label{eq:weak}
    \operatorname{Align}_\text{weak} = s(u_i, v_i) > \max_{j \in N, j\neq i} \{s(u_i, v_j)\}.
\end{equation}

We can then measure the proportion of pairs in the corpus where the property applies.

Strong alignment requires that the nearest neighbours \textit{in general} of a word (or sentence) representation be its translations, \textit{and} therefore that representations of dissimilar source-language words be more distant than the target-language translation.
In terms of the same parallel corpus as above, this can be expressed as:

\begin{equation}\label{eq:strong}
       \operatorname{Align}_\text{strong} = s(u_i, v_i) > \max_{j \in N, j\neq i} \{s(u_i, \boldsymbol{u}_j)\}.
\end{equation}
Note the bolding for emphasis.
These equations refer to bilingual corpora, but can be applied to multiple language pairs in order to measure alignment across a set of languages.

Strong alignment inherently requires greater distance of dissimilar meanings within a language.
That said, weak alignment also benefits from increasing the distance to representations of dissimilar meanings within a language, since accurate retrieval can otherwise be hindered by hubness issues.
The high anisotropy observed in Transformer models \citep{ethayarajh-2019-contextual, gao2018representation} may 
contribute to hubness issues and make it harder to distinguish similar from dissimilar representations.

To see why cross-lingual alignment, particularly ``strong'' alignment, under this view is hard to achieve, consider the \textit{isomorphism assumption} as stated for cross-lingual static embeddings \citep[cf.][]{vulic-etal-2020-good}.
This states that
spaces
of both languages
should have (roughly) the same shape, measured by, for example,
\textit{relational similarity} \citep{vulic-etal-2020-good} or \textit{eigenvector similarity} \citep{sogaard-etal-2018-limitations}.

The isomorphism assumption may not always hold due to cultural-semantic differences, imperfect translation of concepts (e.g., \citealp{gibson-etal-2017-color-names}), typological differences, different corpus domains, different data sizes, and more \citep{ormazabal-etal-2019-analyzing}.
\citet{vulic-etal-2020-good} emphasise that undertraining contributes significantly to non-isomorphism in static embeddings, and this may well apply to contextual models.

We can think of cross-lingual alignment as a complex optimisation problem in this light---to be completely cross-lingually aligned, the model would have to reconcile both large and small differences between many different language spaces.
This may be intractable without also removing valuable contextual and language-specific information.

\paragraph{View II.} A prediction head trained on a source language should be able to find relevant patterns in the representations of a target language, and classify accordingly.
Although it is tempting to think of cross-lingual alignment in terms of simple measures such as cosine similarity, the prediction head works with the full encoder representations as its input, and can (potentially) use subspaces to that effect. 
Given the many constraints on the representations, it is actually very difficult to attain ``cross-lingual alignment'', particularly strong alignment, under View~I.
However, we can consider language-specific subspaces, as well as features pertaining to specific tasks.
For example, some works find subspaces for morphosyntactic aspects \citep{hewitt-manning-2019-structural, Acs_2023_morphosyntactic}, others find directions encoding token frequency \citep{rajaee-pilehvar-2022-isotropy, puccetti-etal-2022-outlier}.

\citet{chang-etal-2022-geometry} separate types of subspaces by how their means and variances differ in different languages.
That is, if both are similar across languages, the axis is language-neutral.
If the means differ between languages and/or variances are very different, the axis is language-sensitive.

The prediction head is usually a linear layer added after the last encoder layer, with a softmax output.
Its weights are learned during fine-tuning.
For cross-lingual transfer to work, the prediction head must place more weight on features that are relevant to the task than on features that are only relevant to the specific language.
Outright reducing the language component in the full encoder representations (as under View~I) works to achieve this goal.
However, under the second view we additionally consider (subspace) projections of the embedding space $\textbf{E}$: 
We conjecture that for any task $T$, there exists a linear projection such that the language component $\textbf{E}_L$ is reduced and task-relevant features $\textbf{E}_T$ are emphasised.
Cross-lingual transfer for the task should succeed if:

\begin{equation}\label{eq:proj}
 \exists \mathrm{Proj(\textbf{E})} \rightarrow |\textbf{E}_T|  > |\textbf{E}_L|
\end{equation}
and if the prediction head is able to find such a projection.
``Strong alignment'' (Equation~\ref{eq:strong}) implies that Equation~\ref{eq:proj} is also true, but
the inverse is not the case.
This second view is particularly salient for fine-tuning tasks, e.g., classification or question answering.

\enote{JL}{I like the definition, but I am not sure about the norm-like notation. At least it deserves a comment what the norm means.}
\enote{KH}{yeah, I'm not totally sure about the norm-like notation. I basically mean ``the size (or saliency) of that component''. could remove the pipes completely maybe?}

\paragraph{In summary,}
the first view of cross-lingual alignment implies a complex optimisation problem (similar to CLWEs), 
as models need to reconcile many variables in order to align representation well.
As we discuss,
this may well be intractable or trade off too much valuable information.
The second view explains why cross-lingual transfer works anyway.
Both views are of course related, and pursue the same overall goals.
Importantly, though, we argue that the first view---focusing on (cosine) similarity between pairs of full vector representations---may be overly simplistic and lose sight of details.

\subsection{Measuring Cross-Lingual Alignment}\label{subsec:measurements}

Cross-lingual alignment or language-neutrality has been measured using a range of metrics, none of which show the full picture:

\paragraph{Cosine similarity} is often used as the similarity function $s$ in Equations~\ref{eq:weak} and~\ref{eq:strong}.
Using the notation from above, the cosine similarity of a translation pair is:

\begin{equation}\label{eq:cosine}
    \mathrm{S_C}(\vec{u_i},\vec{v_i}) = \cos(\theta) = \frac{\vec{u_i} \cdot \vec{v_i}}{\|\vec{u_i}\| \|\vec{v_i}\|}.
\end{equation}

Note that the average cosine similarity in the space can be quite high \citep{ethayarajh-2019-contextual,rajaee-pilehvar-2022-isotropy}, so it is advisable to normalise similarities by the average cosine similarity in order to differentiate values more clearly.

\paragraph{Word or sentence retrieval tasks.} 
Using cosine similarity (Equation~\ref{eq:cosine}), we can then count the proportion of pairs in the bilingual corpus where Equations~\ref{eq:weak} or~\ref{eq:strong} apply as a measure of cross-lingual alignment in the space.
Retrieval tasks are essentially formulated in this way, since they measure whether the closest retrieved element is the correct one.
The tasks may consist only of matched pairs (e.g., Tatoeba), or include ``decoy'' elements that do not have an equivalent (e.g., BUCC2018).
Cosine similarity is commonly used,
or an adjusted retrieval score such as CSLS \citep{lample2018word}.

\paragraph{Zero-shot transfer after fine-tuning,} similarly to retrieval tasks, is both an aim in itself and a proxy for how well-aligned the representations are.
This seems to be the main way that the more complex second view of cross-lingual alignment is translated into metrics.
Interventions before and/or during fine-tuning have been shown to improve transfer performance.
The metrics 
depend on the respective task, but a common way to highlight cross-lingual transfer is to report the \textit{transfer gap}, i.e., the difference between source language performance and the average target language performance.

\paragraph{Language identification} is sometimes used \citep[e.g.,][]{libovicky-etal-2020-language} to measure language-specific elements of the representations.
In this thinking, if a language classifier trained on the output representations performs poorly, then the model representations are highly language-neutral.
This is actually an even stricter goal than ``strong alignment''.
However, it neglects that the representations can have both language-neutral and language-specific areas, and that some language-specific information is necessary.

\paragraph{Visualisation.}
Finally, though not a metric, we mention \textit{t-SNE} \citep{vandermaaten08atsne} here.
This is a visualisation method where spaces are projected down into two or three dimensions for graphing, and it can be extremely helpful to get a better sense of what the space looks like.
However, remember that due to the down-projection and selection of examples, we 
can see only some aspects of the representation space at any given time.

\section{Taxonomy of Alignment Strategies}\label{sec:increase-alignment}

\begin{table*}[!ht]
    \centering
\small
   \renewcommand{\arraystretch}{1.25}
    
    \begin{tabular}{p{2.2cm}|p{7.2cm}|p{5.1cm}}
        \textbf{Objectives}
        & \textbf{From Existing Model} & \textbf{From Scratch} \\
        \hline
        \textbf{Parallel, \mbox{sentence-level}} & \cellcolor{existing-1} Multilingual S-BERT \citep{reimers-gurevych-2020-making}; Sentence-level MoCo \citep{pan-etal-2021-multilingual}; OneAligner \citep{niu-etal-2022-onealigner}; One-pair supervised \citep{tien-steinert-threlkeld-2022-bilingual}; mSimCSE supervised \citep{wang-etal-2022-english}; LaBSE \citep{feng-etal-2022-language}; LAPCA \citep{abulkhanov-etal-2023-lapca}
        & \cellcolor{new-1} LASER \citep{artetxe-schwenk-2019-massively}; LASER3 \citep{heffernan-etal-2022-bitext}; LASER3-CO \citep{tan-etal-2023-multilingual} \\
        
        \textbf{Parallel, \mbox{word-level}} & \cellcolor{existing-2} \citet{cao2020multilingual};
        \citet{wu-dredze-2020-explicit}; 
        Joint-Align + Norm \citep{zhao-etal-2021-inducing};
        VECO \citep{luo-etal-2021-veco}; 
        WEAM \citep{yang-etal-2021-bilingual};  
        XLM-Align \citep{chi-etal-2021-improving};
        WAD-X \citep{ahmat-etal-2023-wadx};
        \citet{efimov-etal-2023-impact}
        & \\
        \textbf{Parallel, \mbox{both levels}} & \cellcolor{existing-3} \citet{kvapilikova-etal-2020-unsupervised}*; InfoXLM \citep{chi-etal-2021-infoxlm}; nmT5 \citep{kale-etal-2021-nmt5}; 
        HiCTL \citep{wei2021on}; \mbox{ERNIE-M} \citep{ouyang-etal-2021-ernie}; DeltaLM \citep{ma2021deltalm}; WordOT \citep{alqahtani-etal-2021-using-optimal};
        & \cellcolor{new-3} ALM \citep{yang-etal-2020-alm}; AMBER \citep{hu-etal-2021-explicit}; XLM-E \citep{chi-etal-2022-xlm}; XY-LENT \citep{patra-etal-2023-beyond} \\
\hdashline
        \textbf{Target~task data} & \cellcolor{existing-4} xTune \citep{zheng-etal-2021-consistency}; FILTER (teacher model) \citep{fang-etal-2021-filter}; XeroAlign \citep{gritta-iacobacci-2021-xeroalign}; Cross-Aligner \citep{gritta-etal-2022-crossaligner}; X-MIXUP \citep{yang2022enhancing} 
        & \cellcolor{new-4} FILTER (student model) \\
        \textbf{Other sources} & \cellcolor{existing-5} RotateAlign \citep{kulshreshtha-etal-2020-cross}; CoSDA-ML \citep{qin-etal-2020-cosda}; DuEAM \citep{goswami-etal-2021-cross}; Syntax-augmentation \citep{ahmad-etal-2021-syntax}; RS-DA \citep{huang-etal-2021-improving-zero}; EPT/APT \citep{ding-etal-2022-simple}; mSimCSE NLI supervision \citep{wang-etal-2022-english}
        & \cellcolor{new-5} DICT-MLM \citep{chaudhary2020dictmlm}; ALIGN-MLM \citep{tang2022alignmlm} \\
        \textbf{Monolingual only} & \cellcolor{existing-6} MAD-X \citep{pfeiffer-etal-2020-mad}; Adversarial \& Cycle \citep{tien-steinert-threlkeld-2022-bilingual}; BAD-X \citep{parovic-etal-2022-bad}; X2S-MA \citep{hammerl-etal-2022-combining}; mSimCSE unsupervised \citep{wang-etal-2022-english}; LSAR \citep{xie-etal-2022-discovering}
        & \cellcolor{new-6} RemBERT \citep{chung2021rethinking}; \xlmr\ XL \& XXL \citep{goyal-etal-2021-larger}; 
        mT5 \citep{xue-etal-2021-mt5}; 
        \mbox{XLM-V} \citep{liang2023xlmv}; mDeBERTaV3 \citep{he2023debertav}; \\

    \end{tabular}
    \caption{Proposed strategies for improved zero-shot transfer by training objectives and initialisation (training from scratch vs. modifying an existing model). 
    *Uses only monolingual data and/or synthetic parallel data.
    }
    \label{tab:data-vs-from-scratch}
\end{table*}

We report on methods for improving zero-shot transfer and increasing cross-lingual alignment.
Table~\ref{tab:data-vs-from-scratch} shows all included papers, organised by initialisation and data requirements of their proposed objectives.
In this section, we discuss each category with examples.
Additionally, we describe some strategies which are applied across different data requirements and initialisations.
We note how methods fit with the two views introduced in \S~\ref{sec:understanding-alignment}.
Some methods are left out here because they are less relevant to the overall analysis, though we explain them in Appendix~\ref{app:more-models} for completeness.

\paragraph{Inclusion of Papers.}

We collected papers for this survey over several search iterations.
We found relevant papers by searching the ACL Anthology, Semantic Scholar, and arXiv.org, as well as following the citation graph.
The initial search terms were ``zero-shot cross-lingual transfer'' and ``cross-lingual alignment''.
We excluded papers where we could not find a PDF version online, and papers focusing on static cross-lingual word embeddings.
We prioritised papers focused on a general notion of cross-lingual alignment over papers applying the concept to a single specific task.

\subsection{Objectives using Parallel Data}\label{subsec:objectives}

First, we discuss models using external parallel data---sentence-parallel or word-parallel.
These make up a plurality of methods in this survey.
In some cases, a sentence-parallel corpus is used and word-level alignments are induced before training.
We tabulate the methods based on whether the proposed objectives focus on word-level alignments, or only sentence-level ones.
``Both levels'' refers mostly to methods using multiple alignment objectives.
In many cases, the alignment objective is combined with a regularisation or joint objective, typically
either masked language modelling (MLM), or minimising the distance from the original model weights (e.g., \citealp{cao2020multilingual}).
In some cases, a newly proposed alignment objective is combined with an existing objective such as translation language modelling (TLM).

\paragraph{Word-level alignment.}

\citet{cao2020multilingual} is an influential early work in explicit cross-lingual alignment training, using
parallel texts. 
The objective is ``contextual word retrieval'', searching for word matches over the entire corpus using
CSLS \citep{lample2018word}, which deals better than cosine similarity with hubness issues.
To regularise the model, they minimise the distance to its initialisation.
This paper clearly takes a similarity-based view of cross-lingual alignment, evaluating on word retrieval tasks but also some languages of XNLI.
A number of works have followed their lead in approaching cross-lingual alignment in this way.
For instance, \citet{wu-dredze-2020-explicit} propose a similar objective with a contrastive loss, which is ``strong'' or ``weak'' based on whether negative examples are considered from both the source and target language or only from the target language.
\citet{zhao-etal-2021-inducing} also use a similar alignment process and combine it with batch normalisation, i.e., forcing ``all embeddings of different languages into a distribution with zero mean and unit variance''.
XLM-Align \citep{chi-etal-2021-improving} combines denoising word alignment with self-labelled word alignment in an EM manner.

\paragraph{Word- and Sentence-level.}

These models either use multiple objectives, or use objectives that are hard to categorise as either word- or sentence-level. 
For instance,
\citet{hu-etal-2021-explicit} propose both a \textit{Sentence Alignment}
and a \textit{Bidirectional Word Alignment} objective inspired by MT for their AMBER model, which they train from scratch.

Among modified models,
\citet{chi-etal-2021-infoxlm} propose the sentence-level cross-lingual momentum contrast objective for InfoXLM.
However, they also emphasise the importance of MLM and TLM (translation language modelling) for token-level mutual information, casting both in information-theoretic terms.

\citet{alqahtani-etal-2021-using-optimal}, meanwhile, 
formulate cross-lingual word alignment as an optimal transport problem.
The mechanism of optimal transport means this is closer to the projection-based view of alignment.
Their input data consists of parallel sentences, but as part of their training process they still focus on finding matched words between the source and target sentences.

\paragraph{Sentence-level alignment.}

Models specifically targeting sentence-level tasks are typically concerned only with sentence-level alignment.
One of these is
multilingual Sentence-BERT \citep{reimers-gurevych-2020-making}, an \xlmr\ model tuned with an English S-BERT model as a teacher.
Using parallel data, the model learns to represent target language sentences similarly to the English source.
This method mostly focuses on similarity scores and achieves good cross-lingual retrieval performance. 
By contrast, LaBSE \citep{feng-etal-2022-language} relies entirely on monolingual data and mined parallel data, but is pre-trained with standard MLM and TLM.
Then, it uses translation ranking with
negative sampling and additive margin softmax \citep{yang-etal-2019-additive} 
to train sentence embeddings.

Among pre-trained models,
LASER \citep{artetxe-schwenk-2019-massively} is a 5-layer BiLSTM trained on machine translation, with the decoder being discarded.
Its successor LASER3 \citep{heffernan-etal-2022-bitext} is a 12-layer Transformer model, but trained using a student-teacher setting, where the teacher is similar to the original LASER.
The follow-up also emphasises lower-resource languages, training a student for each group of similar languages.

\subsection{Contrastive Learning}

Contrastive learning has become popular in NLP for a variety of use cases.
For cross-lingual alignment, it has also been used in several papers, since it aims to increase the similarity of positive examples and the dissimilarity of negative examples jointly.
In effect, contrastive learning should improve hubness issues and increase strong cross-lingual alignment as per the first view in~\S~\ref{subsec:definitions}.

Contrastive learning can be used very effectively on the word level (see InfoXLM, HiCTL, \citet{wu-dredze-2020-explicit}).
For example, HiCTL \citep{wei2021on} stands for Hierarchical Contrastive Learning, which includes both a sentence-level and a word-level contrastive loss.

Nevertheless, contrastive learning is especially popular for sentence embedding models.
Examples include OneAligner \citep{niu-etal-2022-onealigner}, which targets two sentence retrieval tasks, and
is an \xlmr\ version trained on OPUS-100 data.
One version uses all available English-centric pairs, another only uses the single highest-resource corpus, while setting a fixed data budget.
Their training objective uses BERT-Score for similarity scoring, with in-batch normalisation and negatives.

\citet{abulkhanov-etal-2023-lapca}, for their retrieval model LAPCA, also aim for ``strong'' cross-lingual alignment, mining both roughly parallel positive passages and hard negatives.
mSimCSE \citep{wang-etal-2022-english} is another contrastive framework using in-batch negatives, with multiple supervised and unsupervised settings.
LaBSe also uses contrastive learning to achieve good sentence-embeddings, and among pre-trained models, LASER3-CO \citep{tan-etal-2023-multilingual} extends LASER3 by adding contrastive learning to the distillation process.

\subsection{Modified Pre-Training Schemes}

Although many strategies rely on parallel data,
several models are trained from scratch using only monolingual data while modifying specific aspects: a larger vocabulary (XLM-V, \citealp{liang2023xlmv}), rebalanced pre-training vs. fine-tuning parameters (RemBERT, \citealp{chung2021rethinking}).
Several use training objectives that had been tested in an English-only context, such as mDeBERTaV3 \citep{he2023debertav} and mT5 \citep{xue-etal-2021-mt5}.
mDeBERTaV3 additionally uses \textit{gradient-disentangled embeddings}.
Meanwhile, \citet{goyal-etal-2021-larger} significantly scale up model size, producing models with 3.5B and 10.7B parameters.

Like mDeBERTaV3,
XLM-E \citep{chi-etal-2022-xlm} is pre-trained using the ELECTRA training scheme \citep{Clark2020ELECTRA:}, but XLM-E does use both monolingual and parallel data.
The later XY-LENT \citep{patra-etal-2023-beyond} uses the same objectives, but focuses on \textit{many-to-many} bitexts rather than only English-centric data.

\subsection{Adapter Tuning}

Another group of methods use adapters to modify existing models.
MAD-X \citep{pfeiffer-etal-2020-mad} and BAD-X \citep{parovic-etal-2022-bad} are both adapter-based frameworks, combining language adapters and task adapters for improved cross-lingual transfer performance. 
The latter builds on the former by using `bilingual' language adapters, which are trained on monolingual corpora of both the source and the target language.
WAD-X \citep{ahmat-etal-2023-wadx} is another, later method that adds ``word alignment adapters'' using parallel text.

In a somewhat different approach, \citeposs{luo-etal-2021-veco} VECO uses a ``plug-and-play'' cross-attention module 
which is trained during continued pre-training,  
and can be used again in fine-tuning if appropriate parallel data is available.

\subsection{Data Augmentation}

A few methods create pseudo-parallel data by mining sentence pairs or machine translating monolingual text.
For example,
\citet{kvapilikova-etal-2020-unsupervised} fine-tune XLM-100 using TLM, but they do this with 20k synthetic translation pairs, which they create for this purpose.
However, there are also more complex data augmentation strategies being proposed: 
\citeposs{yang-etal-2020-alm} Alternating Language Model (ALM) uses artificially code-switched sentences constructed from real parallel data.
\citet{yang-etal-2021-bilingual} propose a ``cross-lingual word exchange'', where representations from the source language are used to predict target language tokens. 

DICT-MLM \citep{chaudhary2020dictmlm} and ALIGN-MLM \citep{tang2022alignmlm}
both rely on a bilingual dictionary resource.
DICT-MLM trains the model to predict translations of the masked tokens.
ALIGN-MLM rather combines traditional MLM with an alignment loss to optimise average cosine similarity between translation pairs.
CoSDA-ML \citep{qin-etal-2020-cosda} also uses dictionaries in a similar way, but is not trained from scratch.

\subsection{Transformation of Representations}

These methods directly transform the representation space, meaning they lean towards the subspace view of cross-lingual alignment.
They may still be influenced by View I, for example in using bilingual dictionaries.
For instance, RotateAlign \citep{kulshreshtha-etal-2020-cross} uses either dictionaries or parallel data---although parallel data is more effective---to find transformation matrices for each of the last four Transformer layers, combined with language-centering normalisation.

LSAR \citep{xie-etal-2022-discovering} works without any parallel data, by projecting away language-specific elements of the representation space.
The in-batch normalisation used by \citet{zhao-etal-2021-inducing} and \citet{niu-etal-2022-onealigner} is based on the intuition that centering individual language-subspaces will lead to closer cross-lingual alignment.

With the fine-tuning framework X-MIXUP \citep{yang2022enhancing}, the transformation is rather built into the fine-tuning process in a translate-train setting.
It adds MSE between source and target to the fine-tuning loss, as well as the Kullback-Leibler divergence of source and target probability distributions for classification tasks.

\subsection{Tuning with Task Data}

We have so far focused on methods for pre-training or continued pre-training.
Some methods do fine-tuning on the task data and cross-lingual alignment in the same step, often using (translated) task data for a translate-train setting.
Such methods cannot be directly compared to the zero-shot transfer setting, but they are very effective for good transfer performance on the target tasks.

These include xTune \citep{zheng-etal-2021-consistency}, a fine-tuning framework for cross-lingual transfer tasks which can be combined with other models.
xTune also includes \textit{consistency regularisation}, which can work without translated data.
\citeposs{gritta-iacobacci-2021-xeroalign} XeroAlign adds a Mean-Squared-Error (MSE) loss between the source and target sentence to the fine-tuning process. 
Cross-Aligner \citep{gritta-etal-2022-crossaligner} further adds a loss operating on entity level.
\citeposs{fang-etal-2021-filter} FILTER framework first trains a teacher model in the translate-train paradigm, then a student model is trained with a self-teaching loss designed to bridge the gap of label transfer across languages.

\section{What We Do and Don't Know}\label{sec:state-of-the-field}

In this section, we discuss broad findings both from the alignment methods summarised above and from recent related analysis papers.
Future work in this area should follow from open questions.

A selection of task results achieved by the alignment methods can be found in Appendix~\ref{app:eval-models}.
Additionally, Appendix~\ref{app:reprod} shows which authors provide code or model downloads for reproducibility.

\paragraph{Contrastive training works.}
Contrastive training is effective for cross-lingual transfer (as well as for other problems), presumably because it reduces the hubness problem and forces models to differentiate representations.
It is especially popular for retrieval-based tasks and sentence-level models.

\paragraph{Pre-training is not everything.}

Whether a model is newly-trained or modified from a pre-trained model does not appear to determine cross-lingual performance (cf. Appendix~\ref{app:eval-models}).
As models get bigger, only newly pretrained models are available, but at smaller sizes, models \textit{modified} for cross-lingual alignment perform very well.

\enote{AF}{Say what doesn't work and why?}

\paragraph{Related languages are more aligned.}
Closely related languages are more aligned within the models.
In keeping with the many factors encoded by the representations, it makes sense that as more elements differ between languages, they also become more distant in the representation space. Some realignment strategies proved effective at reducing the gap to distant languages, but the pattern remains strong even then.
Accordingly, some papers applying cross-lingual transfer to specific tasks leverage groups of related languages \citep[e.g.,][]{zeng-etal-2023-soft}. 
The many-to-many translation model M2M-100 \citep{fan2021EnglishCentricMultilingualMachine}
similarly makes use of this by grouping training languages.

\paragraph{Use available parallel data.}
While zero-shot cross-lingual transfer is an interesting research problem and can improve pre-trained models, practitioners aiming to deploy the best models should make use of available parallel data.
Models in the translate-train setting consistently outperform those in the zero-shot setting (cf. Appendix~\ref{app:eval-models}), and even a small number of training examples may help.

\paragraph{To what extent is strong alignment necessary?}
In terms of View I, `strong' alignment is often seen as desirable, but we question whether this should be a main goal.
As discussed in \S~\ref{sec:understanding-alignment}, there is a risk of trading off language-specific information.
As for downstream tasks, most cross-lingual retrieval tasks query only the target language space, ignoring the language component.
Thus, strong alignment may not be necessary for these tasks.
LAReQA \citep{roy-etal-2020-lareqa} does test for strong alignment specifically, but it is the exception rather than the rule.
In tasks with a prediction head, strong alignment would ensure that the condition of View~II is met.
Indeed, \citet{gaschi-etal-2023-exploring}
show that strong alignment correlates with better performance on downstream tasks, though they specifically look at classification tasks, not retrieval tasks.
However, this does not tell us if strong alignment is \textit{necessary} for improved transfer performance on these tasks, only that it is correlated.

\paragraph{Try using more source languages.}
There is relatively little work attempting to fine-tune models on two (or more) annotated source languages for zero-shot cross-lingual transfer.
The main paradigm of zero-shot cross-lingual transfer, and a large number of relevant tasks, works with a singular source language (largely English).
However, it stands to reason that learning the target task in two or more annotated source languages would encourage the model to attend more to language-agnostic components as per View II, since this would demonstrate the task as orthogonal to the source language itself.
X-MIXUP and xTune are examples of fine-tuning on multiple languages using the translate-train data in the target task and are found to be quite effective.
However, the translate-train approach does require fine-tuning data in all target languages, which may not always be available.

\section{Multilingual Generative Models}\label{sec:multiling-gen}

Recently, the field has turned much attention to generative Large Language Models (LLMs).
We believe multilingual capabilities of generative models will be more and more important as applications scale.
Thus, we point out several areas of future research, including
how cross-lingual alignment may interact with multi- and cross-lingual generation.

\subsection{Multilinguality As An Afterthought?}

The flagship models \citep[e.g.,][]{touvron2023llama,jiang2023mistral,openai2024gpt4}
in the generative space tend to be English-centric, with some multilingual capability.
However, the proportion of multilingual data is often relatively small and a performance drop-off for mid- and low-resource languages is noticeable \citep[e.g.,][]{ahia-etal-2023-languages}.

Due to the significant costs of pre-training, only few models have been trained to be intentionally multilingual from the start (e.g., BLOOM, \citealp{workshop-2023-bloom}; XGLM, \citealp{lin-etal-2022-shot}).
Unfortunately, even these skew more heavily towards English data than, e.g., XLM-R, and they tend to underperform compared to, e.g., newer GPT, Llama or Mistral models.
mGPT \citep{shliazhko-etal-2024-mgpt} has a more balanced language distribution---the model is comparable with XGLM on few-shot classification tasks, but the evaluation on generative tasks is somewhat limited.

Given this research landscape, we see room to 1) explore why efforts like XGLM or BLOOM have been less effective than retrainings of English-centric models, beyond the oft-cited `Curse of Multilinguality'  \citep{chang-etal-2023-multilinguality}; and 2) be even more intentional about balancing training data, and including multiple languages.
This will necessarily include further data collection efforts as well as additional work on effective multilingual pretraining.

\subsection{Cross-Linguality in Generative Models}

However, numerous efforts have recently extended open models
to new languages.
Examples include EEVE \citep{kim2024EfficientEffectiveVocabulary} for Korean, PolyLM \citep{wei2023polylm} with a focus on Chinese, \citet{andersland2024amharic} for Amharic, or the ongoing Occiglot project \citep{occiglot-2024} for European languages.
Some models, like Tower \citep{alves2024tower} and ALMA \citep{xu2024alma} are trained with a focus on machine translation.
Yet other methods, such as \citet{tanwar-etal-2023-multilingual,huang-etal-2023-languages,zhang2024PLUGLeveragingPivot}, rather focus on prompting strategies to achieve good cross-lingual transfer without necessarily retraining the models.

The Aya Initiative collected instruction data for over 100 languages and varieties, releasing Aya-101 \citep{ustun2024aya}, initialised from mT5, and Aya-23 \citep{aryabumi2024aya}, initialised from Cohere's Command R model.
The latter paper appeals to the `Curse of Multilinguality' for Aya-101 and other massively multilingual models, and focuses on training Aya-23 with those 
languages where a larger amount of data is available. 
By contrast, MaLa-500 \citep{lin2024mala500} continues training from Llama-2 and emphasises training on a very broad set of languages.

\citet{zhang_bayling_2023} tune their model with instruction data and interactive translation examples in order to improve both translation and cross-lingual instruction following.
\citet{li2023align} tune the hidden representations of the first layer using translated data and contrastive learning, which they combine with cross-lingual instruction tuning---this is one of few works on generative models explicitly using a notion of cross-lingual alignment.
\citet{li-murray-2023-zero} use two annotated source languages for cross-lingual generation.

Still, there is plenty of potential for future work in this space.
Cross-lingual alignment of representations has not been explored extensively for generative models.
Importantly, training schemes will need to enable the model to focus on language-neutral \textit{and} language-specific information at the relevant times.
Training for cross-lingual alignment may be combined with sparse fine-tuning approaches such as LoRA \citep{hu2021lora}.
Ideally it should also be integrated by those pre-training new multilingual models.
The goal is to enable better transfer of information between languages
while outputting text in the relevant language.

\subsection{Where to Align a Generative Model?}

Cross-lingual alignment as we discussed it has been researched primarily in encoder-only models, and some encoder-decoder models.
Encoder-only models transform the inputs into a latent space representation which is typically used by a shallow downstream task ``head''. 
For any tasks where the set of outputs does not depend on the language, the model needs to primarily rely on language-neutral axes of the representations.
This is clearly the case for the classification and sequence labelling tasks where encoder models shine.
The task head has somewhat limited opportunity to transform the encoder representations.
Intuitively, strong cross-lingual alignment can be helpful here, including more ``radical'' methods such as mean-centering language-specific embeddings.
We have seen this born out by the proliferation and success of cross-lingual alignment methods in encoders.

In an encoder-decoder framework, there is an explicit delineation with different architectures, where the encoder still outputs a latent space representation, while the decoder predicts the next tokens one-by-one.
Cross-lingual alignment techniques from an encoder context can be naturally applied to the encoder of an encoder-decoder model.
However, there could be risks---``radical'' alignment methods, e.g., mean-centering language-specific embeddings from the encoder, might suppress necessary language-specific information for the decoder, or ``break the link'' of the encoder to the decoder.
Nevertheless, this is certainly a worthwile avenue to explore.

In decoder-only models, there is no such architectural delineation.
Thus, there is no one obvious point at which cross-lingual alignment should be greatest.
Further, the classic zero-shot cross-lingual transfer paradigm may encounter issues in generative settings, since strongly-aligned representations can lead to generation in the wrong language \citep{xue-etal-2021-mt5, li-murray-2023-zero}.

That said, interpretability studies suggest that even in decoder-only models, early layers may perform a kind of encoding role.
For instance, \citet{wendler-etal-2024-llamas} use an early exiting technique known as \textit{logit lens} \citep{nostalgebraist-2020-logit-lens}
to illustrate at which layers in the model translating logits to tokens starts to produce reasonable outputs. 
They demonstrate that Llama-2 ``works'' in English until late in the model, and only then surfaces tokens from the target language.
Meanwhile, \citet{chuang-etal-2024-dola} work only in English but detect hallucinated vs. correct outputs using `differential likelihoods' of early-exited tokens throughout the model.
Both papers suggest inflection points
in mid-to-late layers, where the correct output language or token starts to become more likely.
If we can find a kind of `implicit encoder' in this way, this could become a target for cross-lingual alignment methods.

\subsection{Evaluation of Multilingual Generation}

Benchmarking multilingual generation presents unique challenges compared to both multilingual classification
and monolingual generation tasks.

Several benchmarks have been proposed in the past \citep{asai-etal-2023-buffet, ahuja-etal-2023-mega, gehrmann-etal-2022-gemv2}.
However, these benchmarks 
include a number of reformulated classification tasks in addition to some translation or summarisation tasks.
Classification tasks are not the main strength of generative models, and fine-tuned encoder models often do better there
\citep{lin-etal-2022-shot}.
Translation and summarisation tasks can be evaluated relatively easily using a reference text---which is likely why they were included in the benchmarks.

However, due to the open-ended nature of many generation tasks, even in high-resource languages they are non-trivial to evaluate.
Increasingly, researchers use ChatGPT as a judge on open-ended generation (e.g., \citealp{liu-etal-2023-g}), but this approach 
may not be reproducible due to frequent changes of the ChatGPT model, and
it is less likely to
work well for
low-resource languages.
Note also that care should be taken to avoid leaking evaluation data to proprietary models \citep{balloccu-etal-2024-leak}.

\section{Conclusions}

We 
have surveyed the literature
around cross-lingual alignment, 
providing a taxonomy of methods.
We clarified two main views of the concept, noting how simplistic popular measurements are.
We summarised insights from the surveyed methods and related analysis papers.
Going forward, new challenges present themselves with respect to multilingual generative models: Simply maximising cross-lingual alignment can lead to wrong-language generation.
We thus call for methods which effectively trade-off cross-lingual semantic information with language-specific axes, allowing models to generate fluent and relevant content in many languages.

\section*{Limitations}

\subsection*{Bilingual vs. Multilingual Alignment}\label{subsec:bi-vs-multi}

Since we are talking about highly multilingual models, we are implicitly concerned with multilingual cross-lingual alignment.
However, most of the parallel data involved in (re-)aligning the models or measuring transfer performance are parallel with English.
Thus, in practice, bilingual alignments with English as a pivot language are the most common.
To the extent that alignment in the models is measured (see~\S~\ref{subsec:measurements}), this is typically also done using English as the source language, and less often between a non-English source and non-English target language.
These circumstances significantly limit the training and evaluation of many-to-many cross-lingual alignment.

\subsection*{Multimodality}\label{subsec:multimodal}

Alignment of representations between modalities adds further complexities compared to cross-lingual alignment.
We omit multimodal models from this survey, but note that cross-modal alignment should be similarly examined in future work.

\section*{Acknowledgements}

This publication was supported by LMUexcellent, funded by the Federal Ministry of Education and Research (BMBF) and the Free State of Bavaria under the Excellence Strategy of the Federal Government and the Länder; and by the German Research Foundation (DFG; grant FR 2829/4-1).
The work at CUNI was supported by Charles University project PRIMUS/23/SCI/023.

\bibliography{anthology,custom}
\bibliographystyle{acl_natbib}

\appendix

\section{Further Models Explained}\label{app:more-models}

We add here brief explanations of additional models which we omitted from the main body.
These are methods that did not add as much to the discussion in Section~\ref{sec:increase-alignment}, for example because they are quite similar to other methods.

\subsection{More Word- and Sentence-Level methods}\label{subsec:app-word-sentence}

\citet{pan-etal-2021-multilingual} propose a sentence-level momentum contrast objective, which they combine with TLM to train mBERT.
This seems to be a similar idea to InfoXLM.
nmT5 \citep{kale-etal-2021-nmt5} combines T5 training with a standard MT loss, which arguably targets both granularity levels.

DeltaLM \citep{ma2021deltalm} is also an encoder-decoder model using T5-style training objectives on monolingual and parallel data.
This model is initialised with InfoXLM and modified from there.
\citet{ouyang-etal-2021-ernie} propose the new objectives Cross-Attention MLM and Back-Translation MLM for \mbox{ERNIE-M}.

\subsection{Further Sentence-embedding models}\label{subsec:app-sentence}

\citet{tien-steinert-threlkeld-2022-bilingual} propose two different methods, one supervised by a single language pair not unlike OneAligner, and one unsupervised approach.
Their unsupervised approach uses an adversarial loss encouraging language distributions to become indistinguishable, and a Cycle loss to keep them from degenerating.
In both cases, they freeze the parameters of \xlmr\ and only train a linear mapping.
Their one-pair supervised model is competitive with OneAligner on BUCC2018, but lags further behind on Tatoeba-36, which contains more languages.

\subsection{More Tuning with Task Data}\label{subsec:app-task-data}

DuEAM \citep{goswami-etal-2021-cross} uses data from the XNLI dataset while targeting semantic textual similarity and bitext mining tasks.
The objectives used are Word Mover's Distance and a translation mining loss.
The model performs reasonably well but does not reach the performance of S-BERT.
\citet{efimov-etal-2023-impact} do not tune directly with task data, but they do combine a \citet{cao2020multilingual}-style alignment objective with task fine-tuning in a multi-task-learning objective, with mixed success.

\subsection{More Data Augmentation}\label{subsec:app-data-aug}

RS-DA \citep{huang-etal-2021-improving-zero} is ``randomised smoothing with data augmentation''---a kind of robustness training during fine-tuning, using synonym sets to create the augmented (English) data.
\citet{ding-etal-2022-simple} build on the idea of robust regions and synonym-based data augmentation, adding three objectives to `push' and `pull' the embeddings and attention matrices appropriately (EPT/APT in Table~\ref{tab:finetuned-perf-zs}.
This model performs well on PAWS-X but does not stand out on XNLI.

\subsection{Other Approaches}\label{subsec:app-other}

X2S-MA \citep{hammerl-etal-2022-combining} is an approach using monolingual data to first distill static embeddings from \xlmr, which are then aligned post-hoc and used to train the model for similarity with the aligned static embeddings.
This model takes a similarity-based view and works well on Tatoeba.

\citet{ahmad-etal-2021-syntax}, meanwhile, augment mBERT with syntax information using dependency parses.
They employ a graph attention network to learn the dependencies, which they then mix using further parameters with some attention heads in each layer.

\section{Evaluation of ``Aligned'' Models}\label{app:eval-models}

For reference, we provide a collection of results that the surveyed models achieved on several downstream tasks.
There is no single metric reported by all these papers.
Many report performance on XNLI \citep{conneau-etal-2018-xnli}, in the zero-shot transfer and/or translate-train settings.
A few other tasks are also popular, and we chose a small selection of word- and sentence-level tasks for this overview.
Tables~\ref{tab:finetuned-perf-zs} and~\ref{tab:finetuned-perf-tt} show zero-shot transfer and translate-train results for XNLI, UD-POS \citep{zeman-etal-2019-ud25}, PAWS-X \citep{yang-etal-2019-paws} and MLQA \citep{lewis-etal-2020-mlqa}.

Cross-lingual retrieval is also popular, although the specific tasks reported vary.
We show results for Tatoeba-36 \citep{artetxe-schwenk-2019-massively} and BUCC2018 \citep{ZWEIGENBAUM18.12}, as implemented by \citet{hu2020xtreme}, in Table~\ref{tab:retrieval-perf}.

Unfortunately, there are a number of cases where authors report results for a task but do not use all test languages of the most commonly-used version, meaning that the average results are not comparable.
We omit the results in those cases.

\subsection{What works well?}

First, whether the model is newly-trained or modified from a pre-trained model does not appear to determine performance.
WordOT, a modified mBERT with an optimal transport objective, yields the best result in its size band.
In the next size group, the newly-trained mDeBERTaV3 performs best.
This is a model trained with only monolingual data, with the ELECTRA pre-training objective and additional improvement in the form of gradient-disentangled embeddings.
XLM-E, which does not have this additional element, does markedly worse than mDeBERTaV3.
Only few points behind, ERNIE-M, HiCTL and the JointAlign+Norm method sit at a near-identical performance.
All modify an existing model in different ways: InfoXLM uses information theory, ERNIE-M focuses on aligning the attention parameters, whereas JointAlign+Norm looks at the output vector space.
In the next group, xTune's consistency regulation proves highly effective, with 
ERNIE-M$_{large}$ and InfoXLM just behind.

In both zero-shot transfer and translate-train, once we cross the threshold of 1B parameters, XY-LENT$_{XL}$ is the best available method---we do not know, at this point, if this model would be outperformed by another method being scaled up.
Trained from scratch, XY-LENT specifically uses a lot of parallel data that is not only English-centric, which seems to work well.
XLM-R$_{XXL}$ lags behind XY-LENT$_{XL}$ and XLM-E$_{XL}$ while outperforming its own XL counterpart.
Interestingly, mT5, which underperforms in smaller configurations, is competitive in XL size and does very well in XXL.

In the translate-train setting, mDeBERTaV3 again wins its size group.
However, in the next larger group of models, X-MIXUP proves the most effective.
This method also improves mBERT's performance by a large margin.
X-MIXUP directly addresses representation discrepancies between different languages by linear interpolation between the hidden states of translation pairs.
HiCTL, VECO, and ERNIE-M$_{large}$ come close to the performance of X-MIXUP on this task, while needing more resources.
The contrastive learning approaches in these tables do well (HiCTL, InfoXLM), although they are not necessarily the most performant.
We must add the caveat that not all relevant models are listed in the tables, since not all papers report the full XNLI results.

For Tatoeba, the range of results is especially large---the task has indeed been criticised for its large variability.
Here, contrastive training approaches are both very common and very successful.
LaBSE, OneAligner, and mSimCSE with NLI supervision attain the best overall results.
LaBSE uses both negative sampling and additive margin softmax, OneAligner uses in-batch negatives, and mSimCSE follows a contrastive training approach as well, indicating the strength of these methods for the task.
OneAligner additionally uses in-batch normalisation to offset the hubness problem.

\subsection{What to use?}

Besides the obvious conclusion that larger models usually outperform smaller ones, we recommend using (multi-directional) parallel data if available, and designing models carefully.
Mined or pseudo-parallel data can fulfil that function in some cases.
Use the available translated task data when optimising for a specific application.
When pre-training encoder models,
ELECTRA-style replaced token detection may be the way to go.
Contrastive learning is popular for good reason, especially in the retrieval paradigm.
Methods like OneAligner
also show that models can learn from one language pair to transfer better to multiple language pairs.
Representation normalisation and ensuring that language means are closer together can be very effective and make models competitive with larger ones.
These could also be helpful when not enough data or resources are available for a larger training effort.

Results on zero-shot transfer overall show a similar picture to XNLI, although details change.
For example, XLM-ALIGN's performance stands out on UD-POS but is ``only'' competitive on the other tasks.
HiCTL, meanwhile, is fairly competitive in zero-shot XNLI performance but falls a bit further behind in Table~\ref{tab:finetuned-perf-zs}.
The authors of mDeBERTaV3 do not report any of these other tasks, leaving  XLM-ALIGN, XLM-E$_{base}$, and ERNIE-M$_{base}$ to take the top spots: they all perform well on these three tasks but alternately take the lead.

In the translate-train setting (Table~\ref{tab:finetuned-perf-tt}), VECO$_{in}$ performs best on all three tasks, with HiCTL$_{large}$ on par for PAWS-X but not UD-POS or MLQA.
For XNLI, the best translate-train performance was attained by X-MIXUP, which still does well on these tasks.
Again, overall trends are very similar as for the XNLI task.

Finally, BUCC2018 (Table~\ref{tab:retrieval-perf}) also conveys a similar picture as Tatoeba, although the variation is smaller, likely due the larger datasets and smaller selection of relatively high-resource languages.
mSimCSE with NLI supervision performs best on this task---it also proved effective on \mbox{Tatoeba-36}.
OneAligner is the second most effective on BUCC, with \citeposs{tien-steinert-threlkeld-2022-bilingual} one-pair supervision a close third.

\subsection{Limitations}\label{subsec:eval-tasks}

XNLI is reported in a plurality of papers in our survey, more often than any other single task.
The relative prevalence of XTREME \citep{hu2020xtreme} means that this and several other tasks, including UD-POS, MLQA, PAWS-X, Tatoeba, BUCC2018 and NER, are frequently reported in specific configurations.
Most of these tasks are also popular on their own.
Unfortunately, despite this, many papers do not report results for the full range of ``standard'' target languages, a problem that is more common the more target languages appear in a task.
This particularly limits the ability to compare models across lower-resource languages, and we strongly urge researchers to report results for all standard languages when evaluating on a task.

\begin{table*}[t]
    \centering
    \begin{tabular}{l|ccccc}
        \textbf{Model} & \textbf{Size} & \textbf{XNLI} & \textbf{UDPOS} & \textbf{PAWS-X} & \textbf{MLQA (F1)}\\
        \hline
        \textit{Zero-shot transfer} \\
        \hline
         mBERT \citep{hu2020xtreme} & 110M & 65.4 & 71.5 & 81.9 & 61.4 \\
         \rowcolor{existing-5} mBERT + Syntax augm. & $\sim$110M & -- & -- & 84.3 & 60.3 \\
         \rowcolor{existing-5} mBERT + EPT/APT & $\sim$110M & 68.4 & -- & \textbf{86.2} & -- \\
         \rowcolor{new-5} DICT-MLM & $\sim$110M & 68.6 & 71.6 & 84.8 & -- \\
         \rowcolor{existing-2} mBERT+JointAlign+Norm & $\sim$110M & 72.3 & -- & -- & -- \\
         \rowcolor{existing-2} WordOT & $\sim$110M & \textbf{75.4} & -- & -- & -- \\
         \rowcolor{new-3} AMBER & 172M & 71.6 & -- & -- & -- \\
         \hdashline
         \rowcolor{existing-5} XLM-R$_{base}$ + EPT/APT & $\sim$270M & 75.8 & -- & 87.1 & -- \\
         \rowcolor{existing-2} XLM-ALIGN & $\sim$270M & 76.2 & \textbf{76.0} & 86.8 & 68.1 \\
         \rowcolor{existing-3} InfoXLM$_{base}$ & $\sim$270M & 76.5 & -- & -- & 68.1 \\
         \rowcolor{existing-3} ERNIE-M$_{base}$ & $\sim$270M & 77.3 & -- & -- & \textbf{68.7} \\
         \rowcolor{existing-3} HiCTL$_{base}$ & $\sim$270M & 77.3 & 71.4 & 84.5 & 65.8 \\
         \rowcolor{existing-2} XLM-R+JointAlign+Norm & $\sim$270M & 77.6 & -- & -- & -- \\
         \rowcolor{new-6} mDeBERTaV3 & $\sim$276M & \textbf{79.8} & -- & -- & -- \\
         \rowcolor{new-3} XLM-E$_{base}$ & 279M & 76.6 & 75.6 & \textbf{88.3} & 68.3 \\
         \rowcolor{new-6} mT5$_{small}$ & 300M & 67.5 & -- & 82.4 & 54.6 \\
         \hdashline
         \rowcolor{new-3} XY-LENT$_{base}$ & 447M & 80.5 & -- & 89.7 & 71.3 \\
         XLM-R \citep{hu2020xtreme} & 550M & 68.2 & 73.8 & 86.4 & 71.6 \\
         \rowcolor{existing-3} HiCTL$_{large}$ & $\sim$550M & 81.0 & 74.8 & 87.5 & 72.8 \\
         \rowcolor{existing-3} InfoXLM$_{large}$ & $\sim$550M & 81.4 & -- & -- & 73.6 \\
         \rowcolor{existing-3} ERNIE-M$_{large}$ & $\sim$550M & 82.0 & -- & 89.5 & 73.7 \\
         \rowcolor{existing-4} XLM-R$_{large}$ + xTune & 550M & \textbf{82.6} & \textbf{78.5} & \textbf{89.8} & \textbf{74.4}\\
         \rowcolor{new-6} RemBERT & 575M & 80.8 & 76.5 & 87.5 & 73.1 \\
         \rowcolor{new-6} mT5$_{base}$ & 580M & 75.4 & -- & 86.4 & 64.6 \\
         \rowcolor{existing-2} VECO$_{out}$ & 662M & 79.9 & 75.1 & 88.7 & 71.7 \\
         \rowcolor{new-6} XLM-V & $\sim$750M & 76.0 & -- & -- & 66.0 \\
         \rowcolor{new-3} XLM-E$_{large}$ & 840M & 81.3 & -- & -- & -- \\
         \hdashline
         \rowcolor{new-3} XY-LENT$_{XL}$ & 2.1B & \textbf{84.8} & -- & -- & -- \\
         \rowcolor{new-3} XLM-E$_{XL}$ & 2.2B & 83.7 & -- & -- & -- \\
         \rowcolor{new-6} XLM-R$_{XL}$ & 3.5B & 82.3 & -- & -- & 73.4 \\
         \rowcolor{new-6} mT5$_{XL}$ & 3.7B & 82.9 & -- & 89.6 & 73.5 \\
         \hdashline
         \rowcolor{new-6} XLM-R$_{XXL}$ & 10.7B &  83.1 & -- & -- & 74.8 \\
         \rowcolor{new-6} mT5$_{XXL}$ & 13B & \textbf{85.0} & -- & 90.0 & \textbf{76.0} \\
    \end{tabular}
    \caption{Zero-shot transfer XNLI performance reported by various papers, ordered by model size.
    Many papers do not report exact parameter counts, so we make an estimate ($\sim$) based on the model they modify, or on hyperparameters where reported.
    We draw dashed lines between models of markedly different sizes.
    }
    \label{tab:finetuned-perf-zs}
\end{table*}


\begin{table*}[t]
    \centering
    \begin{tabular}{l|ccccc}
        \textbf{Model} & \textbf{Size} & \textbf{XNLI} & \textbf{UDPOS} & \textbf{PAWS-X} & \textbf{MLQA (F1)} \\
       \hline
        \textit{Translate-train} \\
        \hline
         mBERT \citep{hu2020xtreme} & 110M & 74.6 & -- & 86.3 & 65.6 \\
         \rowcolor{existing-4} mBERT + X-MIXUP & 110M & \textbf{78.8} & 76.5 & \textbf{89.7} & \textbf{69.0} \\
         \hdashline
         \rowcolor{existing-3} InfoXLM$_{base}$ & $\sim$270M & 80.0 & -- & -- & -- \\
         \rowcolor{existing-3} ERNIE-M$_{base}$ & $\sim$270M & 80.6 & -- & -- & -- \\
         \rowcolor{new-6} mDeBERTaV3 & $\sim$276M & \textbf{82.2} & -- & -- & -- \\
         \rowcolor{new-6} mT5$_{small}$ & 300M & 72.0 & -- & 79.9 & 64.3 \\
         \hdashline
         \rowcolor{new-3} XY-LENT$_{base}$ & 447M & 82.9 & -- & 92.4 & -- \\
         \rowcolor{existing-4} XLM-R$_{large}$ + xTune & $\sim$550M & 82.6 & 78.5 & 89.8 & 75.0 \\
         \rowcolor{existing-4} FILTER & $\sim$550M &  83.6 & 76.2 & 91.2 & 75.8 \\
         \rowcolor{new-4} FILTER + Self-teaching & $\sim$550M & 83.9 & 76.9 & 91.5 & 76.2 \\
         \rowcolor{existing-3} ERNIE-M$_{large}$ & $\sim$550M & 84.2 & -- & 91.8 & -- \\
         \rowcolor{existing-3} HiCTL$_{large}$ & $\sim$550M & 84.5 & 76.8 & \textbf{92.8} & 74.4 \\ 
         \rowcolor{existing-4} XLM-R$_{large}$ + X-MIXUP & 550M & \textbf{85.3} & 78.4 & 91.8 & 76.5 \\ 
         \rowcolor{new-6} mT5$_{base}$ & 580M & 79.8 & -- & 89.3 & 75.3 \\
         \rowcolor{existing-2} VECO$_{in}$ & 662M & 84.3 & \textbf{79.8} & \textbf{92.8} & \textbf{77.5} \\
         \hdashline
         \rowcolor{new-3} XY-LENT$_{XL}$ & 2.1B & \textbf{87.1} & -- & \textbf{92.6} & -- \\
         \rowcolor{new-6} XLM-R$_{XL}$ & 3.5B & 85.4 & -- & -- & -- \\
         \rowcolor{new-6} mT5$_{XL}$ & 3.7B & 85.3 & -- & 91.0 & 75.1 \\
         \hdashline
         \rowcolor{new-6} XLM-R$_{XXL}$ & 10.7B &  86.0 & -- & -- & -- \\
         \rowcolor{new-6} mT5$_{XXL}$ & 13B & \textbf{87.1} & -- & 91.5 & 76.9 \\
    \end{tabular}
    \caption{Translate-train XNLI, UD-POS, PAWS-X, and MLQA performance reported by various papers, ordered by model size.
    Many papers do not report exact parameter counts, so we make an estimate based on the model they modify, or based on hyperparameters where reported.
    We mark the estimates with a tilde ($\sim$).
    We draw dashed lines between models of markedly different sizes.
    }
    \label{tab:finetuned-perf-tt}
\end{table*}


\begin{table*}[ht]
    \centering
    \begin{tabular}{l|ccc}
        \textbf{Model} & \textbf{Size} & \textbf{Tatoeba} & \textbf{BUCC} \\
        \hline
        mBERT \citep{hu2020xtreme} & 110M & 38.7 & 56.7 \\
        \rowcolor{existing-6} mBERT + LSAR & $\sim$110M & 44.6 & -- \\
         \rowcolor{new-5} DICT-MLM & $\sim$110M & 47.3 & -- \\
         \rowcolor{existing-1} LaBSe & $\sim$110M & \textbf{95.0} & \textbf{89.7} \\
         \hdashline
         \rowcolor{existing-6} X2S-MA & $\sim$270M & \textbf{68.1} & -- \\
         \rowcolor{existing-1} LAPCA-LM$_{base}$ & $\sim$270M & -- & 71.3 \\
         \rowcolor{new-3} XLM-E$_{base}$ & 279M & 65.0 & -- \\
         \hdashline
         XLM-R \citep{hu2020xtreme} & 550M & 57.3 & 66.0 \\
         \rowcolor{existing-3} HiCTL$_{large}$ & $\sim$550M & 59.7 & 68.4 \\
         \rowcolor{existing-6} XLM-R + LSAR & $\sim$550M & 65.1 & -- \\
         \rowcolor{existing-6} T\&ST (unsup) & $\sim$550M & 74.2 & 82.4 \\
         \rowcolor{existing-1} T\&ST (one-pair) & $\sim$550M  & 80.4 & 89.6 \\
         \rowcolor{existing-3} ERNIE-M$_{large}$ & $\sim$550M & 87.9 & -- \\
         \rowcolor{existing-1} LAPCA-LM$_{large}$ & $\sim$550M & -- & 83.5 \\
         \rowcolor{existing-1} OneAligner & 550M & \textbf{92.9} & 90.5 \\
         \rowcolor{existing-6} mSimCSE uns. & $\sim$550M & 78.0 & 87.5 \\
         \rowcolor{existing-1} mSimCSE sup. & $\sim$550M & 88.3 & 88.8 \\
         \rowcolor{existing-5}mSimCSE NLI & $\sim$550M & 91.4 & \textbf{95.2} \\
         \rowcolor{existing-3} \citet{kvapilikova-etal-2020-unsupervised} & $\sim$570M -- &  & 75.8 \\
         \rowcolor{existing-2} VECO$_{out}$ & 662M & 75.1 & 85.0 \\ 
    \end{tabular}
    \caption{Tatoeba-36
    and BUCC 
    performance reported by various papers, ordered by parameter counts.
    }
    \label{tab:retrieval-perf}
\end{table*}

\section{Reproducibility}\label{app:reprod}

In order to reproduce a method, or apply it to a new use case, detailed instructions and ease of reuse are vital.
Providing implementation code is the most straightforward way to ensure that \textit{all} necessary details are conveyed to a reader, and they do not waste time reimplementing them.
Similarly, model downloads save time and make further experimentation much easier.
The larger the model in question, the more important model downloads become, since re-training them requires more time, effort, and compute.

In Table~\ref{tab:code-models-online}, we list all papers that provide their code, a model download, or both.
Some of these are well documented, some not so much.
Some are well-maintained, some not at all.
We did not test the provided code and links, simply checked that they are online and contain what looks to be the promised artifacts.
Papers where we did not find any artifacts are omitted.

\begin{table*}
\renewcommand{\arraystretch}{1.05}
\centering
    \begin{tabular}{l|ccc}
        \textbf{Model Name} & \textbf{Code Available} & \textbf{Model Download} \\
        \hline
         Syntax Augmented mBERT \citep{ahmad-etal-2021-syntax} & yes & no \\
         LASER \citep{artetxe-schwenk-2019-massively} & yes & yes, fairseq \\
         XLM-Align \citep{chi-etal-2021-improving} & yes & yes, HF \\
         InfoXLM \citep{chi-etal-2021-infoxlm} & yes & yes, HF \\
         RemBERT \citep{chung2021rethinking} & no & yes, HF \\
         EPT/APT \citep{ding-etal-2022-simple} & yes & no \\
         \citet{efimov-etal-2023-impact} & no & no \\
         FILTER \citep{fang-etal-2021-filter} & yes & no \\
         LaBSE \citep{feng-etal-2022-language} & no & yes, TFH, HF \\
         X2S-MA \citep{hammerl-etal-2022-combining} & yes & no \\
         XLM-R$_{XL/XXL}$ \citep{goyal-etal-2021-larger} & yes & yes, fairseq, HF \\
         XeroAlign \citep{gritta-iacobacci-2021-xeroalign} & yes & no \\
         CrossAligner \citep{gritta-etal-2022-crossaligner} & yes & no \\
         mDeBERTaV3 \citep{he2023debertav} & yes & yes, HF \\
         LASER3 \citep{heffernan-etal-2022-bitext} & yes & yes, fairseq \\
         XLM-V \citep{liang2023xlmv} & no & yes, HF \\
         XGLM \citep{lin-etal-2022-shot} & no & yes, fairseq, HF \\
         VECO \citep{luo-etal-2021-veco} & no* & yes, fairseq \\
         ERNIE-M \citep{ouyang-etal-2021-ernie} & yes & yes, HF \\
         BAD-X \citep{parovic-etal-2022-bad} & yes & yes, AdapterHub \\
         MAD-X \citep{pfeiffer-etal-2020-mad} & no & yes, AdapterHub \\
         Multilingual S-BERT \citep{reimers-gurevych-2020-making} & yes & yes, HF \\
         ALIGN-MLM \citep{tang2022alignmlm} & yes & no \\
         \citet{tien-steinert-threlkeld-2022-bilingual} & yes & no \\
         mSimCSE \citep{wang-etal-2022-english} & yes & yes, HF \\
         \citet{wu-dredze-2020-explicit} & yes & no \\
         LSAR \citep{xie-etal-2022-discovering} & yes & no \\
         mT5 \citep{xue-etal-2021-mt5} & yes & yes, custom, HF \\
         X-MIXUP \citep{yang2022enhancing} & yes & no \\
         JointAlign + Norm \citep{zhao-etal-2021-inducing} & yes & yes \\
         xTune \citep{zheng-etal-2021-consistency} & yes & no \\
    \end{tabular}
    \caption{A list of those surveyed papers that provide code and/or model downloads. 
    We do not test the provided code, only making sure it remains online at time of writing.
    We sort by first author last name. 
    *VECO has a repository online that includes only fine-tuning code.
}
\label{tab:code-models-online}
\end{table*}

\end{document}